\documentclass[paper=a4,11pt,DIV=11,parskip=half]{scrartcl}

\usepackage[utf8]{inputenc}
\usepackage{amsmath,amssymb,amsthm}
\usepackage{graphicx}
\usepackage[ruled,vlined,linesnumbered]{algorithm2e}
\usepackage[breakable]{tcolorbox}
\usepackage{float}
\usepackage{hyperref}
\usepackage[style=vancouver,backend=biber]{biblatex}
\usepackage[margin=2.5cm]{geometry}
\usepackage{titling}
\usepackage{caption}
\everydisplay{\abovedisplayskip=8pt \belowdisplayskip=8pt \abovedisplayshortskip=8pt \belowdisplayshortskip=8pt}

\hypersetup{
    colorlinks=true,
    urlcolor=blue,
    linkcolor=blue,
    citecolor=blue
}

\title{Controllable protein design with particle-based Feynman--Kac steering}

\author{
    Erik Hartman\textsuperscript{1,*} \and 
    Jonas Wallin\textsuperscript{2} \and
    Johan Malmström\textsuperscript{1} \and
    Jimmy Olsson\textsuperscript{3}
}

\begin{document}

\maketitle
\date{}

\vspace{-2em}
\noindent\textsuperscript{1}Division of Infection Medicine, Faculty of Medicine, Lund University, Sweden\\
\textsuperscript{2}Department of Statistics, Lund University, Sweden\\
\textsuperscript{3}Department of Mathematics, KTH, Sweden\\
\textsuperscript{*}Corresponding author: Erik Hartman, \href{mailto:erik.hartman@med.lu.se}{erik.hartman@med.lu.se}

\begin{abstract}     
    Proteins underpin most biological function, and the ability to design them with tailored structures and properties is central to advances in biotechnology. Diffusion-based generative models have emerged as powerful tools for protein design, but steering them toward proteins with specified properties remains challenging. The Feynman–Kac (FK) framework provides a principled way to guide diffusion models using user-defined rewards. In this paper, we enable FK-based steering of RFdiffusion through the development of guiding potentials that leverage ProteinMPNN and structural relaxation to guide the diffusion process towards desired properties. We show that steering can be used to consistently improve predicted interface energetics and increase binder designability by $89.5\%$. Together, these results establish that diffusion-based protein design can be effectively steered toward arbitrary, non-differentiable objectives, providing a model-independent framework for controllable protein generation.
\end{abstract}

\section{Introduction}

Diffusion models have enabled \textit{de novo} protein design by learning to iteratively transform random noise into structured molecular configurations \cite{rfd,se3,diff_wu,trippe_2022,chroma,salad,boltzgen,protpardelle,rfab,rfhelix,jointdiff}. Trained on large structural datasets, these models learn to generate realistic protein backbones with diverse structural and functional properties to, for example, successfully design enzymes \cite{rfd,rf2_enzyme}, antibodies \cite{rfab}, toxin inhibitors \cite{snake}, vaccine components \cite{npv}, and peptide therapeutics \cite{rfpeptides}. While unguided diffusion can produce high-quality designs, the broader utility of these models depends on their ability to generate proteins with specific properties, such as molecular binding, stability, or sequence composition.

The most common approach to steering diffusion-based generation is to condition the model on specified context, such as a target structure and hotspots for binder design \cite{rfd,cfg,rfbeta}. However, this approach requires the conditioning information to be specified at training time. Fine-tuning is an alternative, in which the model is retrained to generate samples with desired properties \cite{discriminator,sag,pg}. However, this must be repeated for each new objective, is typically infeasible for external users, and is computationally expensive. Modifying the Langevin dynamics by applying a gradient-based control term derived from a differentiable reward function is another approach \cite{sde}, but this requires the reward to be differentiable and can be computationally expensive.

These challenges highlight a central limitation of current diffusion-based protein design: the lack of a general inference-time mechanism for steering generation toward specific properties. Given the broad range of potential design objectives in protein engineering, ranging from binding affinity to protein stability to sequence-level constraints, flexible and controllable generation toward different properties is critical for improving generative models in this domain. These limitations motivate the need for a general framework that enables controllable protein generation without retraining or architectural modifications. Such a framework would enable diffusion-based design to be steered toward arbitrary properties, including non-differentiable and black-box objectives, while preserving design diversity and quality.

The Feynman–Kac (FK) framework provides a principled approach to defining tilted distributions by reweighting diffusion trajectories according to potential functions derived from user-defined rewards \cite{gfr,fkc,DelMoral2004,Chopin2020,twisting2,smc_diff,twisting}. Sequential Monte Carlo methods can then be used to sample from these tilted distributions, enabling inference-time steering without altering the underlying generative model. In principle, this allows diffusion dynamics to be guided at inference time by rewards encoding arbitrary design objectives. However, adapting FK-based steering to protein design presents unique challenges as protein properties often require evaluation on all-atom refined structures, necessitating specialized potential functions. Moreover, these potentials are inherently non-differentiable due to stochastic sequence generation and discrete optimization steps, precluding gradient-based approaches.

Here, we develop such guiding potentials and implement a steering framework for RFdiffusion \cite{rfd} using the FK formulation, demonstrating that this approach can steer the generative process toward desired properties. We further show that steering can enhance the designability of generated binders, increasing the likelihood of producing functionally relevant designs. Our implementation is available as a user-friendly Python package on \href{https://github.com/ErikHartman/FK-RFdiffusion}{GitHub}.

\section{Main}

To introduce inference-time control into diffusion-based protein generation, we develop and apply a particle-based implementation of the FK (Feynman–Kac) framework for RFdiffusion. This approach frames protein design as a guided stochastic process in which diffusion trajectories are weighted and resampled according to user-defined rewards, biasing generation toward high-reward configurations. Below, we first outline the steering approach, and then demonstrate how to construct guiding potentials for different design objectives. Lastly, we apply steering to protein binder design, showing that it improves the designability of generated binders across multiple targets.

\subsection{Particle-based steering of protein diffusion models}

Protein diffusion frames protein generation as a discrete denoising process that transforms random noise into structured molecular configurations. Each intermediate state is a protein conformation, where residues are described by their C$\alpha$ positions together with local orientation frames defined by the N--C$\alpha$--C atoms \cite{rfd,se3,diff_wu,trippe_2022}. Generation proceeds by starting from a highly corrupted noise-like structure and iteratively applying a learned reverse transition that gradually restores protein-like geometry, ultimately producing a valid backbone.

Rather than producing a single denoising trajectory, we propagate an \emph{ensemble} of trajectories (particles) in parallel. At selected denoising steps, each particle is assigned a \emph{potential} based on a user-defined reward that reflects how well its current trajectory aligns with the design objective. The particle set is then resampled so that higher-reward trajectories are more likely to be kept and duplicated, while low-reward trajectories are more likely to be discarded. The surviving particles are subsequently propagated by one additional denoising step using the same underlying diffusion model. Repeating this resample--propagate update progressively biases the ensemble toward trajectories that score well under the reward, while retaining stochasticity through both the diffusion transitions and the use of multiple particles (\textbf{Fig.~1a}).

Instead of drawing final designs purely from the model’s native distribution, steering skews sampling toward designs that achieve higher reward under the chosen objective (\textbf{Fig.~1b}). Importantly, this control mechanism does not require retraining, fine-tuning, or architectural modifications of the diffusion model. It is therefore model-independent and compatible with arbitrary objectives, including non-differentiable scores and black-box evaluations.

A practical challenge is that most meaningful objectives such as binding energetics, secondary structure composition, sequence-level properties, cannot be evaluated directly on noisy intermediate diffusion states, which typically lack physically interpretable geometry and do not define an amino-acid sequence. We therefore evaluate rewards on a denoised proxy of the final structure predicted from the current diffusion state by the denoising network. To make this proxy suitable for biochemical scoring, we generate an accompanying sequence using ProteinMPNN \cite{mpnn} and then perform sidechain packing and local all-atom relaxation in PyRosetta \cite{rosetta}. The resulting refined sequence--structure pair is used for property computation and reward evaluation (\textbf{Fig.~1c}; see \textbf{Methods: Rewards}).

\begin{figure}[H]
\centering
\includegraphics[width=\textwidth]{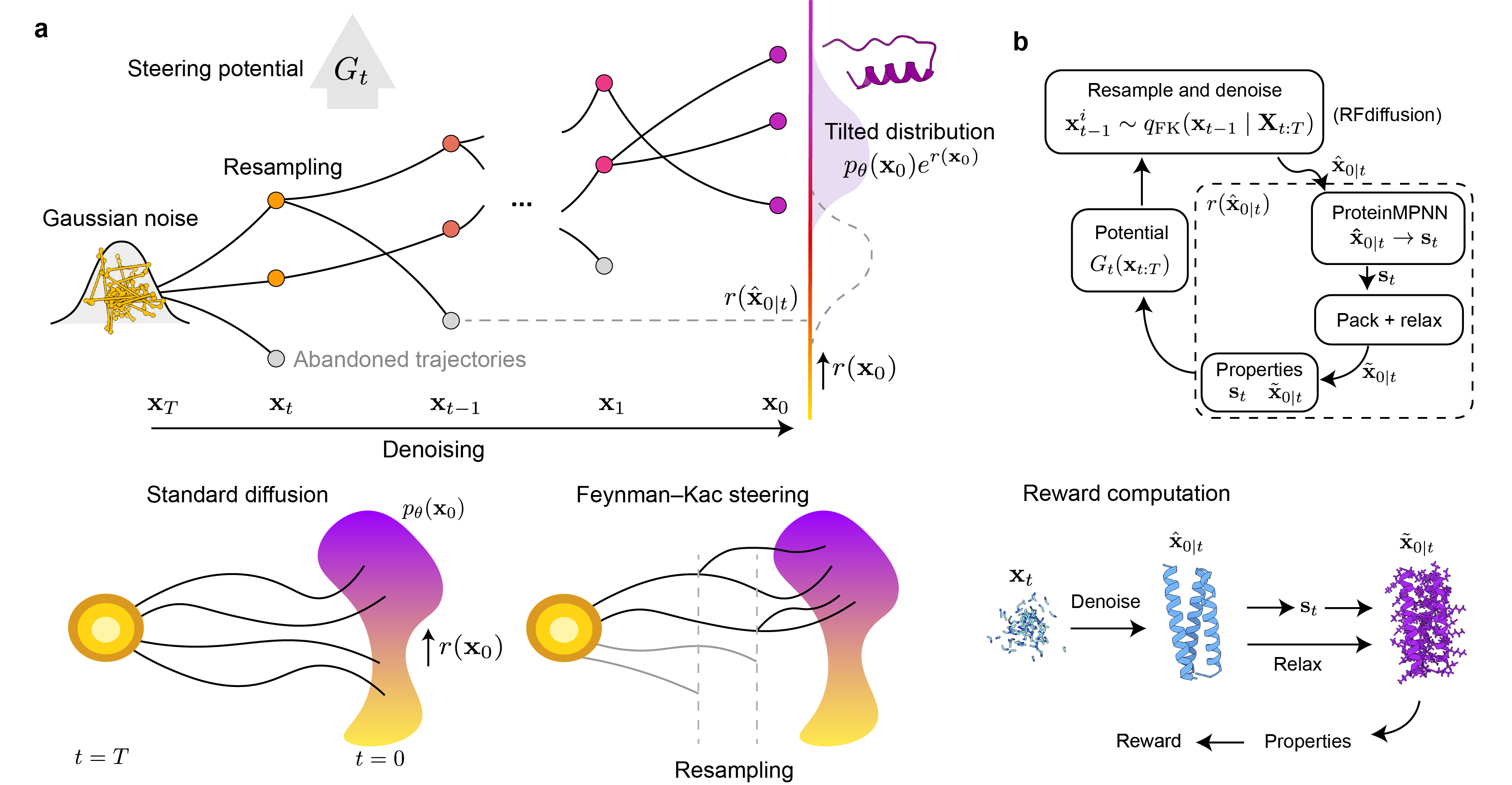}
\caption{\textbf{Particle-based steering of protein diffusion models.}
\textbf{a} Starting from Gaussian noise at diffusion time $T$, multiple diffusion trajectories, $\mathbf{x}_t$, are propagated and resampled at each denoising step according to \textit{steering potentials} $G_t$, which tilt sampling toward high-reward configurations. Rewards are computed on the predicted denoised proxy $\hat{\mathbf{x}}_{0\mid t}$ which is attained by a forward pass through the denoising network. Low-reward trajectories (gray) are likely to be discarded, and are progressively abandoned, while high-reward trajectories are likely to be kept and further denoised toward $\mathbf{x}_0$. This process yields samples from a \textit{tilted distribution} proportional to $p_{\theta}(\mathbf{x}_0)e^{r(\mathbf{x}_0)}$, where $r(\mathbf{x}_0)$ defines the reward function and $p_{\theta}(\mathbf{x}_0)$ is the distribution of proteins that can be designed by the diffusion model. \textbf{b} RFdiffusion is used to denoise $\mathbf{x}_t$ and the predicted denoised proxy $\hat{\mathbf{x}}_{0\mid t}$ is fed to the reward function. In the reward function, the denoised proxy is used to sample an accompanying sequence, $\mathbf{s}_t$ via ProteinMPNN, and the structure is refined with packing and relaxation, followed by property computation, and lastly, potential computation.}
\end{figure} 

To illustrate the flexibility of this framework, we first apply steering to two qualitatively different objectives: a sequence-level reward that controls net charge (\textbf{Fig.~2a}) and a structure-level reward that biases secondary structure composition (\textbf{Fig.~2b}). In both cases, the guided sampling procedure shifts the generated designs toward the desired property while preserving diversity relative to the unguided diffusion baseline.

The most widely applied use of protein diffusion models is binder design, where the goal is to generate peptides or proteins that adopt geometries compatible with a specified target surface and achieve favorable interface energetics. By steering toward a reward that reflects favorable binding, such as $-\Delta G$, we can bias the diffusion process toward trajectories predicted to form stronger interactions with the target protein. This produces a consistent upward shift in interface quality over the course of denoising compared to unguided diffusion, demonstrating that steering can improve binding-oriented objectives without retraining the generator (\textbf{Fig.~2c}).

Finally, we show that steering can also be used to control secondary structure composition in binder designs (\textbf{Fig.~2d}). By applying the same secondary structure reward as in \textbf{Fig.~2b} during binder design, we can bias the generated binders toward specific structural motifs, such as $\beta$-sheets or $\alpha$-helices.

Together, these examples show that steering provides a general inference-time mechanism for controlling diffusion-based protein generation across diverse design tasks, with the same resample--propagate procedure differing only in the choice of reward.

\begin{figure}[H]
\centering
\includegraphics[width=\textwidth]{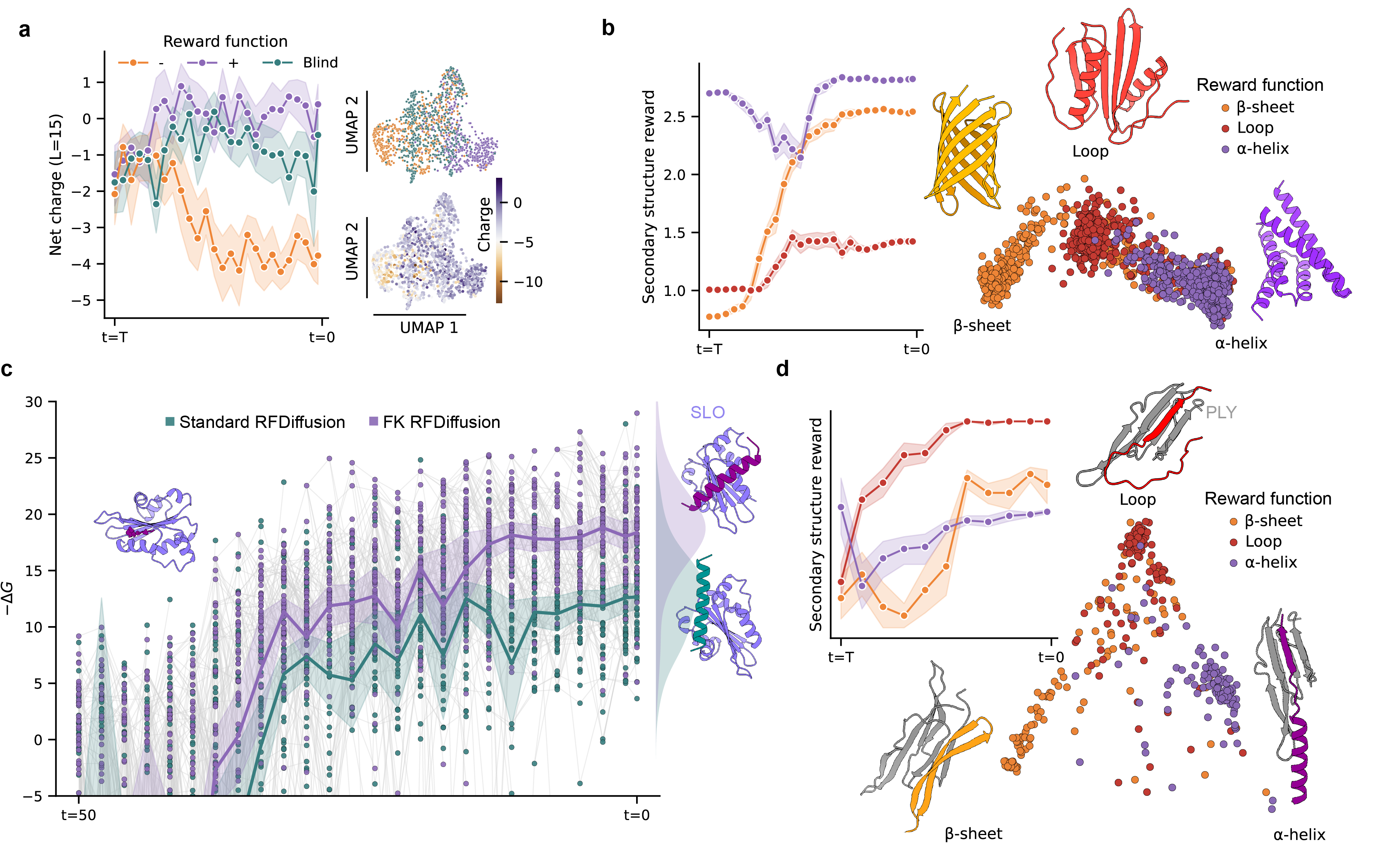}
\caption{\textbf{Steering controls diverse properties across design tasks.}
\textbf{a} Steering by sequence charge reward functions. Lines show the mean net charge over denoising steps for designs guided toward negative charge (orange), positive charge (purple), or unguided (teal) trajectories. Right: UMAP embeddings of generated sequences colored by reward function (top) and charge (bottom). \textbf{b} Steering secondary structure formation by sampling toward $\beta$-sheet (orange), loop (red), or $\alpha$-helix (purple) secondary structures. Left: mean secondary structure reward progression over denoising steps. Right: representative examples of resulting structures, along with a triplot where each point is positioned according to its fractional composition of $\beta$-sheet, loop, and $\alpha$-helix secondary structures. \textbf{c} Comparison of peptide binder design trajectories generated using standard RFdiffusion (teal) and FK-steered RFdiffusion (purple) against streptolysin O (PDB: 4HSC). The y-axis shows $-\Delta G$ (higher is better) across denoising steps. Each dot represents a particle at the given timepoint, with lines connecting its immediate predecessors and successors. \textbf{d} FK steering enables control over secondary structure composition in binder designs. Lines indicate the mean secondary structure reward for designs guided toward $\beta$-sheet (orange), loop (red), or $\alpha$-helix (purple) motifs. Right: representative designed binders and a triplot where each point is positioned according to its fractional composition of $\beta$-sheet, loop, and $\alpha$-helix content. In lineplots, shaded bands indicate the mean ±1 standard deviation.}

\end{figure} 

\subsection{Dynamics of steering and control/diversity trade-offs}

There are multiple ways to formulate the steering potential that determines how rewards influence trajectory resampling during steering (see \textbf{Methods: Feynman–Kac formulation} for details). In brief, potentials can be defined to depend on the current state only (\textit{immediate}), the difference in reward between successive steps (\textit{difference}), the maximum reward over all future steps (\textit{max}), or the cumulative reward over all future steps (\textit{sum}).

These potential formulations shape the trade-off between reward maximization and sequence diversity in distinct ways. To evaluate their impact, we designed binders under identical conditions while varying the potential functions. Strong guidance, especially under \textit{sum}, \textit{max} and \textit{immediate}, yields high terminal rewards (\textbf{Fig. 3a}) but quickly reduces sequence diversity, whereas the \textit{difference} potential maintains a steadier balance between exploration and exploitation (\textbf{Fig. 3b}). The UMAP projection of generated sequences shows that guided samples remain largely within the distribution of the unguided model, consistent with guidance acting as a form of importance weighting rather than fully re-training the generative prior (\textbf{Fig. 3c}). Sequence diversity was quantified as one minus the mean pairwise identity across generated sequences.

Varying steering hyperparameters further modulates this balance: lowering the sampling temperature ($\tau$) enhances rewards but reduces diversity, while resampling less frequently preserves exploration at the cost of control. Increasing the number of particles improved both reward and diversity, and delaying the onset of guidance until the diffusion exited the high-noise regime led to higher final rewards, likely because early rewards are unreliable and can drive premature convergence (\textbf{Fig. 3d}).

\begin{figure}[H]
\centering
\includegraphics[width=\textwidth]{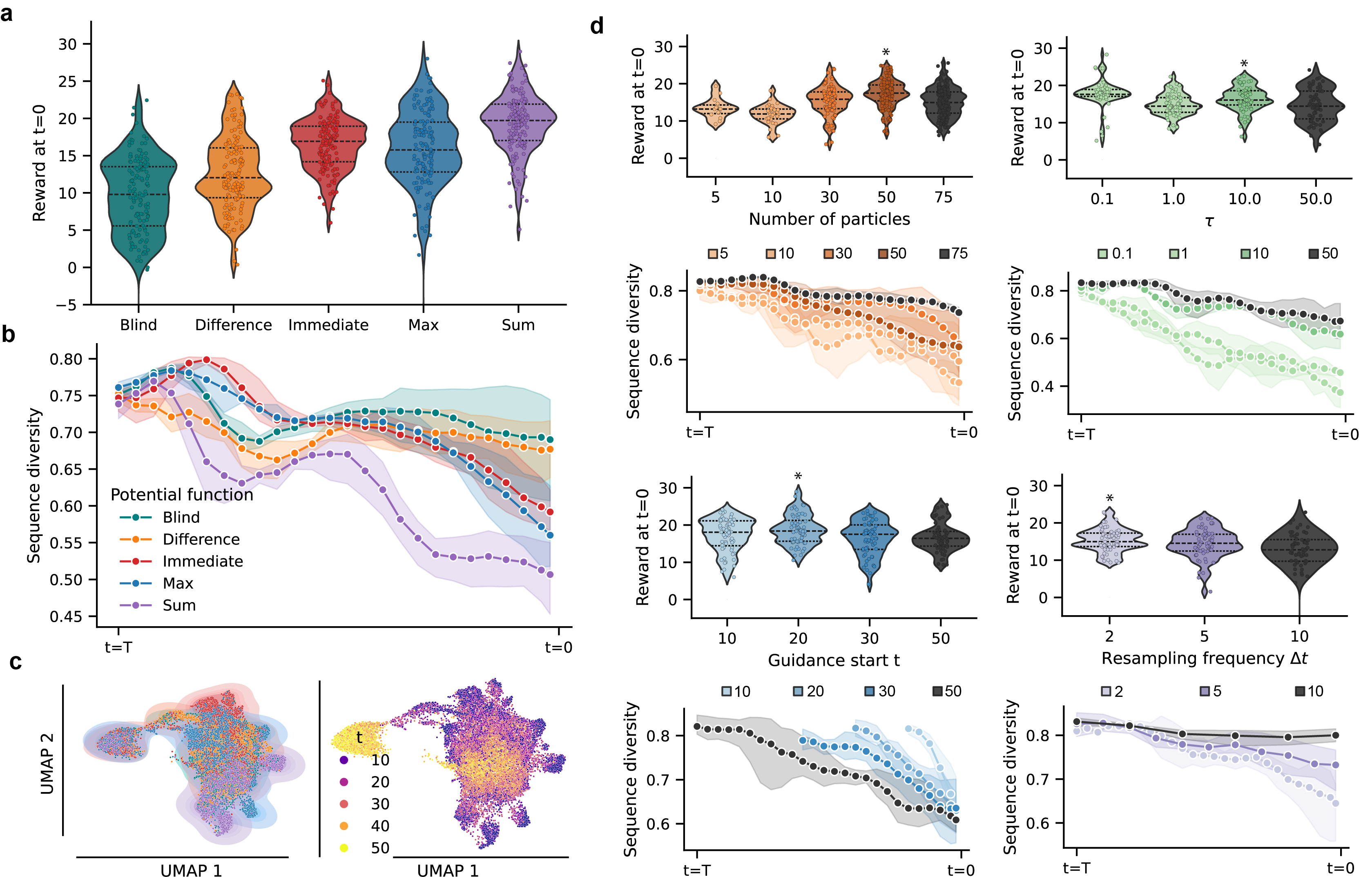}
\caption{\textbf{Effect of steering potentials and parameters on FK-guided binder design.}
\textbf{a} Comparison of terminal rewards across different potential formulations. Violin plots show the reward distribution at $t=0$ for 3 rounds of unguided (blind) diffusion and four steering potentials. \textbf{b} Sequence diversity trajectories for each potential. Lines indicate mean sequence diversity over denoising steps. \textbf{c} UMAP visualization of generated sequences colored by potential type (left) and denoising step (right). \textbf{d} Sensitivity of steering to key hyperparameters. Top to bottom, left to right: reward and sequence diversity at $t=0$ as a function of (i) number of particles, (ii) guidance temperature $\tau$, (iii) guidance start step $t$, and (iv) resampling frequency $\Delta t$. Increasing particle count or lowering $\tau$ improves mean reward but reduces sequence diversity. Streptolysin O (PDB: 4HSC) was consistently used as the target. The parameters deemed optimal and which were later used for binder design are highlighted with a star ($*$).}
\end{figure}

Because the reward function depends on stochastic sequence generation by ProteinMPNN (see \textbf{Methods: Rewards} for details), we examined how repeated sequence sampling influenced the performance of steering (\textbf{Fig. 4a}). At each timestep, multiple sequence–structure pairs were generated for each trajectory and averaged to obtain a more reliable estimate of the reward. Increasing the number of sampled pairs per step consistently raised the mean terminal reward (\textbf{Fig. 4b}) yet left population diversity largely unchanged (\textbf{Fig. 4c}). This demonstrates that more robust reward estimation improves the performance of steering without affecting population diversity, in contrast to earlier experiments where higher rewards often came at the cost of reduced diversity (\textbf{Fig. 3d}). The standard deviation of rewards decreased along the denoising trajectory (\textbf{Fig. 4d}), with the largest uncertainty occurring before $t = 20$. This is consistent with the observed benefit of delaying guidance until later diffusion steps (\textbf{Fig. 3d}), when denoised structures provide sufficient geometric and sequence context for reliable evaluation.

\begin{figure}[H]
\centering
\includegraphics[width=\textwidth]{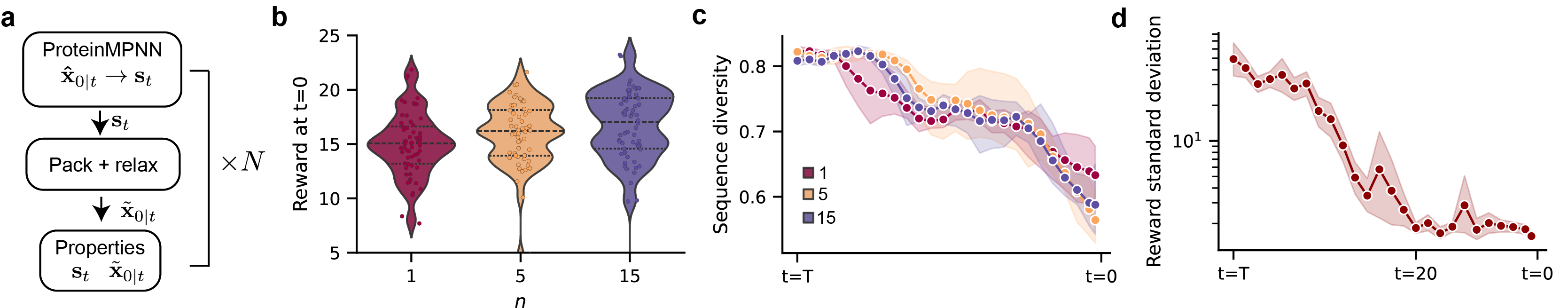}
\caption{\textbf{Effect of repeated reward evaluations on steering.}
\textbf{a} At every time step $t$, $N$ independent evaluations are generated by applying ProteinMPNN followed by sidechain packing and relaxation, and lastly property evaluation. The used reward function is then formed as the average of the reward across these evaluations. \textbf{b} Distribution of rewards at $t=0$ as a function of the number of repeated evaluations. \textbf{c} Sequence diversity trajectories over denoising steps for different repetition counts. \textbf{d} Standard deviation of the rewards across repeated runs as a function of diffusion time. Variability decreases as diffusion progresses toward $t=0$. Triplicate runs were used for evaluation. Shaded bands indicate the mean ±1 standard deviation.}
\end{figure}

\subsection{Guiding toward better interface energetics improves designability of binders}

A key metric in protein design is \textit{designability}, typically defined as the proportion of designed sequences that are predicted to fold into the intended structure and achieve favorable properties such as binding affinity (\textbf{Fig. 5a}). Designability is a critical predictor of experimental success, as designs with higher predicted structural agreement to their targets are more likely to yield functional binders \cite{Bennett2023,Kim2025}. Improving designability is therefore a central goal in binder design, however, direct steering toward improved designability is unfeasible due to the computational cost of structural prediction. However, we reasoned that there may be a correlation between steerable rewards and designability. To investigate this, we generated binders without steering against 5 targets, before re-predicting the complex conformation. We then correlated steerable rewards with designability. We found that the interface free energy reward, which is computationally efficient to evaluate, was strongly correlated with designability across all targets (\textbf{Fig. 5c}). This suggests that steering toward improved interface energetics may provide an effective proxy for improving designability, even without directly optimizing for it (\textbf{Fig. 5d}). 

We then designed binders using steering with the interface free energy reward, and compared their designability to binders generated without steering. Steering improved designability by $> 89.5\%$, demonstrating that guiding toward improved interface energetics can substantially enhance the likelihood of generating structurally coherent and functionally relevant binders (\textbf{Fig. 5e,f}). Further, for structures that passed the designability filter, the interface energy upon re-prediction was markedly improved by steering, indicating that the method not only increases the proportion of designs that are structurally coherent but also enhances their predicted binding affinity (\textbf{Fig. 5g}).

\begin{figure}[H]
\centering
\includegraphics[width=\textwidth]{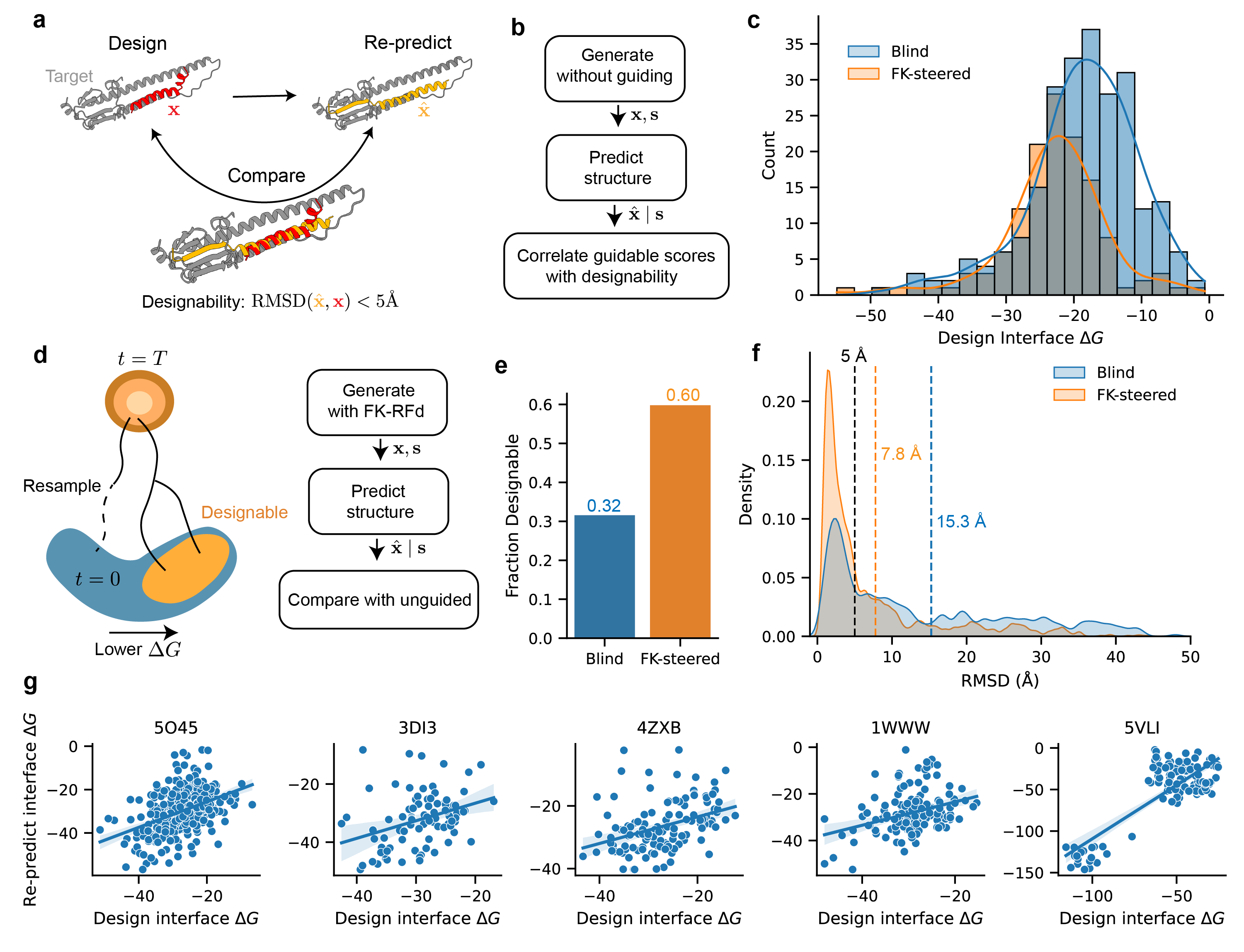}
\caption{\textbf{Guiding toward better interface energetics improves designability of binders.} \textbf{a} Designability is defined as the proportion of generated designs that are predicted to fold into the intended structure ($\text{RMSD} < 5 \text{Å}$) and achieve favorable properties such as binding affinity. For binder design, RMSD was computed over the designed binder only. \textbf{b} Workflow for evaluating the correlation between steerable rewards and designability. \textbf{c} Histogram of interface $\Delta G$ stratified on designability. \textbf{d} Schematic illustrating how steering toward improved interface energetics can enhance designability, and workflow for evaluating the impact of steering on designability. \textbf{e} Comparison of designability between binders generated with and without steering respectively. \textbf{f} Kernel density estimate of RMSD of backbones generated with and without steering respectively. The black dashed line shows the 5 Å cutoff which is often used to define designability. The orange and blue dashed lines show the mean RMSD for the designs generated with and without steering respectively. \textbf{g} Correlation of the interface energy during design and upon re-prediction for designs that passed the designability filter.}
\end{figure}

\subsection{Steering improves reward across diverse bacterial virulence targets}

To assess the generality of steering for binder design, we applied the approach to a panel of clinically relevant virulence factors from important Gram-positive bacterial pathogens, \textit{Streptococcus pyogenes} and \textit{Streptococcus pneumoniae}, including pneumolysin (PLY), streptolysin O (SLO), EndoS, IdeS, and C5a peptidase. These proteins are key pathogenic virulence factors with characterized catalytic or antibody-binding regions which suggests that they accommodate binding. However, to our knowledge, they have not previously been explored as design targets for \textit{de novo} protein binders and are generally structurally diverse. Steering was applied using the interface free energy reward and was run using the optimal parameters identified during benchmarking. Across all targets, steering increased the binding reward compared to unguided diffusion, indicating that this approach improves predicted interface affinity across diverse targets (\textbf{Fig. 6}).

\begin{figure}[H]
\centering
\includegraphics[width=\textwidth]{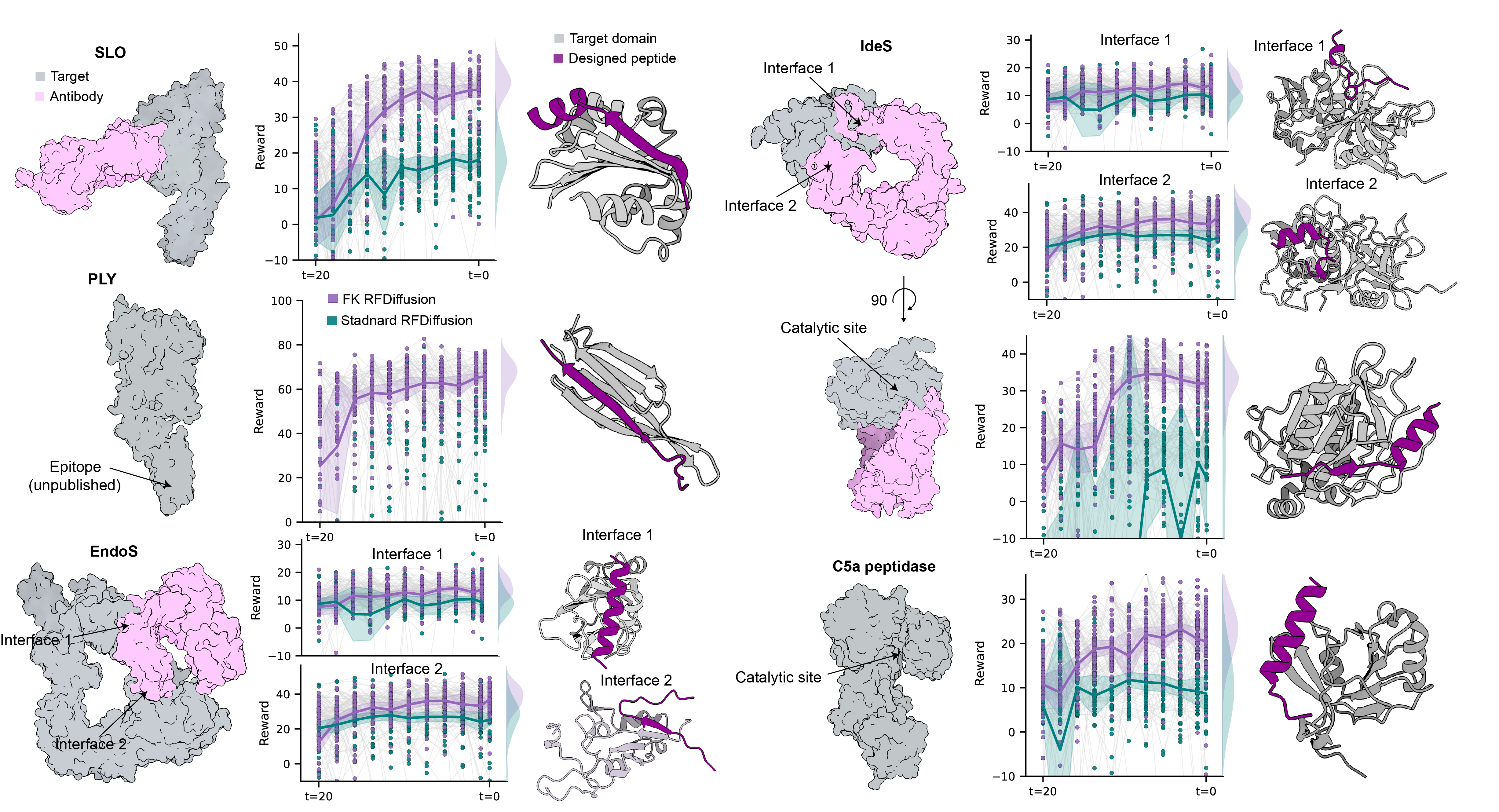}
\caption{\textbf{Steering improves binder design across diverse bacterial virulence targets.}
Steering applied to the \textit{Streptococcus} virulence factors streptolysin O (SLO), pneumolysin (PLY), EndoS, IdeS, and C5a peptidase. For each target, the left panels show the receptor (grey) and its known antibody (pink) or active-site region, which define the design interface. Middle panels show reward trajectories over diffusion timesteps comparing standard RFdiffusion (teal) and steered RFdiffusion (purple). Each dot represents a particle at the given timepoint, with lines connecting its immediate predecessors and successors. Shaded bands indicate the mean ±1 standard deviation. Right panels show representative steered-designed peptides (purple) bound to their corresponding target domains (grey).}
\end{figure}

Together, these results demonstrate that steering provides an effective and general mechanism for introducing controllability into diffusion-based protein generation. Across multiple property classes and biological targets, steering enhances model performance without modifying or retraining the underlying diffusion network, establishing it as a flexible inference-time framework for directed protein design.

\section{Discussion}

Particle-based steering provides a principled means to control diffusion-based protein design without retraining or modifying the generative model. By reweighting trajectories according to user-defined rewards, steering enables targeted exploration of structural space while maintaining diversity. When applied to RFdiffusion and coupled with sequence recovery and relaxation, we show that the FK formulation effectively directs diffusion models toward desired biochemical and structural properties, including charge distribution, secondary structure composition, and predicted binding affinity. The method operates entirely at inference time and supports flexible, property-specific rewards defined over both structure and sequence.

Importantly, we show that guiding toward improved interface energetics improved the designability of binders substantially, demonstrating that steering can enhance the likelihood of generating structurally coherent and functionally relevant designs. We reason that this may be due to the fact that improved interface energetics are more physically plausible and structurally coherent configurations, which are more likely to be predicted as designable by structure prediction-models. 

A central challenge in applying steering to protein design lies in defining a potential function that consistently guides the diffusion process toward regions of structural space associated with desired molecular properties. The first difficulty lies in recovering a denoised structure from a noisy sample, which is fundamentally approximate since many plausible structures can correspond to the same noisy representation. The latter represents a broader limitation of protein design itself, where accurate prediction of properties such as binding affinity or stability remains inherently difficult. In this work, we address these challenges by using the diffusion network to predict the denoised structure, followed by sequence generation with ProteinMPNN and structural relaxation in PyRosetta to obtain physically coherent sequence–structure pairs suitable for evaluation. Both steps alter the potential landscape and impact steering performance. Continued improvements in denoising models and algorithms, and in biophysical property predictors, are therefore likely to enhance the accuracy and stability of steering.

Another practical consideration concerns the selection and optimization of steering parameters. The performance of steering depends on maintaining an appropriate balance between exploration and exploitation, which is influenced by the choice of potential as well as by parameters such as resampling frequency and the time of guidance onset. In this study, we characterized the general behaviour of steering across these parameters, but the optimal configuration is likely dependent on the specific reward formulation and design objective. Developing adaptive schemes that automatically tune these parameters may further improve robustness and generality across different design objectives.

Despite these considerations, steering remains computationally efficient, as it operates entirely at inference time and requires no gradient-based optimization or retraining of the generative model. The method adds only a minor overhead relative to standard sampling, making it practical for large-scale or iterative design tasks. Moreover, its modular formulation allows straightforward extension to new reward functions, enabling users to incorporate additional biochemical or structural objectives without modification of the underlying framework. This flexibility makes steering well suited for integration with emerging predictive models and experimental feedback, providing a general foundation for controllable protein generation.

\section{Methods}

\subsection{Feynman–Kac formulation}
We model protein generation as a discrete denoising diffusion process that transforms random noise into structured molecular configurations. The process consists of two components: a forward diffusion process that progressively adds noise to native protein structures, and a reverse process that learns to recover structure from noise. Each protein conformation $\mathbf{x}_t$ is represented in a rigid-frame coordinate system, where residues are described by their C$\alpha$ positions together with local orientation frames defined by the N-C$\alpha$-C atoms \cite{rfd,se3,diff_wu,trippe_2022}. The forward process defines a sequence of latent variables $\mathbf{x}_0, \mathbf{x}_1, \dots, \mathbf{x}_T$, where $\mathbf{x}_0$ represents the native protein backbone and $\mathbf{x}_T$ approaches a standard Gaussian distribution through successive noising steps $q(\mathbf{x}_{t+1} \mid \mathbf{x}_t)$. The reverse, or generative, process is parameterized by a neural network $p_\theta$ with parameters $\theta$ that iteratively denoises $\mathbf{x}_t$ to generate artificial data $\mathbf{x}_0$ according to
$$
p_{\theta}(\mathbf{x}_{0:T}) = p(\mathbf{x}_T) \prod_{t=0}^{T-1} p_{\theta}(\mathbf{x}_t \mid \mathbf{x}_{t+1}),
$$
where $p(\mathbf{x}_T)$ denotes the distribution over noisy structures. During generation, the model begins from $\mathbf{x}_T \sim p(\mathbf{x}_T)$ and successively generates $\mathbf{x}_0$, corresponding to a valid protein backbone.

To introduce control over the generative process, we use the FK (Feynman–Kac) framework to define a tilted distribution, which we sample using sequential Monte Carlo \cite{DelMoral2004,Chopin2020}. In this framework, an ensemble of particles, each representing a diffusion trajectory $\mathbf{x}_{0:T}$, is propagated and resampled according to the FK potentials. During generation, particles are reweighted by scalar potentials that quantify the extent to which the intermediate structure at step $t$ satisfies a specified objective, steering the ensemble toward high-reward regions of structural space.

The objective of particle-based steering is to reweight the diffusion process so that samples are drawn from a \textit{tilted distribution}
$$
p^\ast(\mathbf{x}_0) = \frac{1}{Z} \int p_{\theta}(\mathbf{x}_{0:T}) e^{r(\mathbf{x}_0)}d\mathbf{x}_{1:T},
$$
where $Z$ is a normalizing constant and $r(\mathbf{x}_0)$ is a reward function. To construct a sequential approximation, we define stepwise potentials $G_t(\mathbf{x}_{t:T})$ such that
$$
\prod_{t=0}^{T} G_t(\mathbf{x}_{t:T}) = e^{r(\mathbf{x}_0)}.
$$
This allows us to express the tilted target distribution $p^*(\mathbf{x}_0)$ as the time-zero marginal of the \textit{tilted path measure}
$$
p^\ast(\mathbf{x}_{0:T})
\propto p_{\theta}(\mathbf{x}_{0:T})\prod_{t=0}^{T} G_t(\mathbf{x}_{t:T}),
$$
which defines the FK-steered diffusion process. During generation, we maintain an ensemble $\mathbf{X}_{t:T} = \big\{{\mathbf{x}_{t:T}^1,\dots,\mathbf{x}_{t:T}^N}\big\}$ of trajectories, referred to as particles, $\mathbf{x}^i_{t:T}$ denoting the $i$th particle. At each reverse step $t$, sampling proceeds through a resampling–propagation update. More precisely, the marginal distribution of $\mathbf{x}_{t-1}$ given the current ensemble $\mathbf{X}_{t:T}$ is
$$
q_{\text{FK}}(\mathbf{x}_{t-1}\mid \mathbf{X}_{t:T})= \sum_j w(\mathbf{x}_{t:T}^j)p_{\theta}(\mathbf{x}_{t-1}\mid\mathbf{x}_t^j),
$$
where the weights $w(\mathbf{x}_{t:T}^j) = G_t(\mathbf{x}_{t:T}^j) / \sum_iG_t(\mathbf{x}_{t:T}^i)$ reflect each trajectory’s normalized potential. Algorithmically, each iteration consists of resampling the current particles, with replacement, according to their weights. This duplicates high-potential particles while eliminating low-potential ones, after which each resampled particle is propagated by extending it with a draw from the reverse transition $p_{\theta}(\mathbf{x}_{t-1}\mid\mathbf{x}_t)$. This sequential Monte Carlo update tilts the ensemble toward high-reward trajectories and enables controllable protein generation.

Evaluating relevant rewards such as geometries or energies directly on noisy diffusion states $\mathbf{x}_t$ is infeasible, as intermediate structures lack meaningful geometry or sequence context. We therefore estimate rewards on a denoised proxy of the terminal structure, $\hat{\mathbf{x}}_{0 \mid t} = f_{\theta}(\mathbf{x}_t,t)$, where $f_{\theta}$ denotes the denoising network which is part of the trained diffusion model and predicts the final structure $\hat{\mathbf{x}}_{0\mid t}$ from the noisy intermediate state $\mathbf{x}_t$. To obtain a physically coherent and sequence-specific model suitable for reward evaluation, $\hat{\mathbf{x}}_{0 \mid t}$ is passed through ProteinMPNN \cite{mpnn} to generate a sequence $\mathbf{s}_t$, followed by sidechain packing and local relaxation in PyRosetta to produce a refined structure $\tilde{\mathbf{x}}_{0 \mid t}$ \cite{rosetta}. The resulting pair $(\tilde{\mathbf{x}}_{0 \mid t}, \mathbf{s}_t)$ defines the refined state on which the reward function operates and we define these refinement steps as part of the reward evaluation pipeline (see \textbf{Methods: Rewards}).

The choice of potential $G_t$ determines how rewards influence trajectory resampling during the diffusion process. In practice, different functional forms of $G_t$ yield distinct trade-offs between responsiveness and stability.

\begin{table}[H]
\centering
\caption{\textbf{Steering potential domain, form and boundary potentials.} $r_t$ is used as shorthand for $r_t(\mathbf{x}_t)$.}
\begin{tabular}{llll}
\hline
Potential & Domain & Functional form & Boundary potentials \\
\hline
Immediate & $\mathbf{x}_t$ & $e^{r_t}$ & $G_0=e^{r_0}\big/\prod_{t=1}^{T}G_t$ \\
Difference & $(\mathbf{x}_t,\mathbf{x}_{t+1})$ & $e^{r_t-r_{t+1}}$ & $G_T=1$ \\
Max & $\mathbf{x}_{t:T}$ & $e^{\max_{s\ge t}r_s}$ & $G_0=e^{r_0}\big/\prod_{t=1}^{T}G_t$ \\
Sum & $\mathbf{x}_{t:T}$ & $e^{\sum_{s=t}^{T}r_s}$ & $G_0=e^{r_0}\big/\prod_{t=1}^{T}G_t$ \\
\hline
\end{tabular}
\end{table}

Among these, the \textit{difference} potential satisfies the exact FK decomposition, while the \textit{immediate}, \textit{max}, and \textit{sum} variants serve as heuristic alternatives that distribute the reward differently along the diffusion trajectory.

For all formulations, the reward includes a guidance scale $\lambda = 1/\tau$,
$$
r_t(\mathbf{x}_t) := \lambda r(\hat{\mathbf{x}}_{0 \mid t}),
$$
where $\hat{\mathbf{x}}_{0 \mid t} = f_\theta(\mathbf{x}_t, t)$ is the denoised structural estimate of the terminal state $\mathbf{x}_0$. The function $r(\cdot)$ operates on the refined sequence–structure pair derived from $\hat{\mathbf{x}}_{0 \mid t}$ as described above, such that $r_0(\mathbf{x}_0) = r(\mathbf{x}_0)$. The scale $\tau$ controls the sharpness of the resampling distribution: lower $\tau$ (higher $\lambda$) enforces stronger selection toward high-reward configurations but reduces particle diversity, whereas higher $\tau$ produces broader, more exploratory ensembles.

\subsection{Implementation}

Particle-based steering was implemented as a wrapper around RFdiffusion \cite{rfd} (commit e220924) and uses its pretrained denoising network ($p_{\theta}$) and diffusion schedule to generate reverse trajectories. The implementation introduces particle-based guidance, where an ensemble of $n_{\text{particles}}$ particle trajectories evolve through reweighting and resampling operations every $\Delta t$ timestep (starting at $t_{\text{start}}$) according to user-defined reward functions.

At each guidance step, a reward function $r_t(\mathbf{x}_{t})$ is evaluated for every particle, and potentials $G_t$ are computed according to the chosen formulation. To maintain numerical stability, exponentiation is performed after subtracting the maximum value and clipping ($r_t - r_{max} \ge -10^3$). Particle weights are then assigned as normalized potentials
$$
w(\mathbf{x}_{t:T}^j) = \frac{G_t(\mathbf{x}_{t:T}^j)}{\sum_iG_t(\mathbf{x}_{t:T}^i)}
$$
Guided diffusion proceeds from the initial noisy state toward $t=1$, alternating between denoising by $p_{\theta}$ and resampling by the FK update.

The computational cost of steering scales linearly with the number of diffusion particles and reward evaluations. Each guidance step requires $n_{\text{particles}}$ forward passes through RFdiffusion and corresponding reward computations, resulting in a total complexity of $\mathcal{O}\big(n_{\text{particles}} \cdot (C_{p_{\theta}} + C_{r})\big)$ where $C_{p_{\theta}}$ is the cost of inference of the denoising network and $C_{r}$ the cost of computing the reward. In practice, diffusion steps are efficiently parallelized across GPU devices, while reward computations, particularly those involving ProteinMPNN inference and PyRosetta relaxation, can be executed in parallel on CPUs, substantially reducing wall time.

\subsection{Sequence recovery and packing}

To obtain a physically meaningful estimate of $\mathbf{x}_0$ for reward evaluation, sequence design from ProteinMPNN (commit 8907e66) \cite{mpnn} was integrated with all-atom refinement in PyRosetta as part of the reward pipeline. This procedure transforms a coarse, denoised backbone representation into a fully atomic, sequence-specific structure better suited for downstream evaluation.

Let ${\mathbf{x}}_{0} \in \mathbb{R}^{3L}$ denote the residue-frame backbone coordinates, and let $\mathcal{A}^L$ be the discrete amino acid sequence space of length $L$. ProteinMPNN defines a conditional distribution $P_{\text{MPNN}}(\mathbf{s}\mid {\mathbf{x}}_{0})$, from which sequences $\mathbf{s}$ are sampled at a temperature of 0.2 using solubility-optimized model weights. Each sequence $\mathbf{s}$ is threaded onto the backbone to generate a full-atom model $\mathbf{x}_{\text{atom}}({\mathbf{x}}_{0}, \mathbf{s}) \in \mathbb{R}^{3N}$, where $N$ is the number of heavy atoms.

We define the transformation $(\mathbf{x}_0, \mathbf{s}) \mapsto (\tilde{\mathbf{x}}_0, \mathbf{s}),$ which represents the combined process of sampling a sequence and reconstructing the corresponding all-atom model through sidechain packing and local energy minimization. The refined coordinates $\tilde{\mathbf{x}}_{0}$ are obtained by discrete rotamer sampling followed by continuous minimization of torsional angles $\chi_1, \chi_2, \dots$ under the Rosetta energy function, resolving steric clashes and optimizing the side chain poses. This defines a conditional distribution
$$
(\tilde{\mathbf{x}}_{0}, \mathbf{s}) \sim P_{\text{ref}}(\cdot \mid {\mathbf{x}}_{0}),
$$
which maps each denoised backbone to an ensemble of relaxed, sequence-specific structures. Reward evaluation is thus performed on physically and chemically coherent sequence–structure pairs.

\subsection{Rewards}

In the FK framework, the reward function is a deterministic function. In our case, evaluating such a reward requires intermediate sampling and refinement steps that introduce randomness. Specifically, the combination of ProteinMPNN sequence sampling and PyRosetta relaxation defines a conditional distribution $P_{\text{ref}}(\cdot \mid \mathbf{x}_0)$, from which realizations $(\tilde{\mathbf{x}}_0, \mathbf{s})$ are drawn. These steps are best regarded as part of the FK update mechanism rather than as intrinsic randomness in the reward function itself. However, for clarity, we treat the entire procedure as an evaluation pipeline that provides an approximation to a deterministic reward defined on the expected refined structure, allowing us to tailor the reward pipeline based on the objective.

To obtain a stable estimate of the expected reward, each denoised backbone ${\mathbf{x}}_{0}$ is evaluated multiple times. Each realization
$$
(\tilde{\mathbf{x}}^{(i)}_{0}, \mathbf{s}^{(i)}_t) \sim P_{\text{ref}}(\cdot \mid {\mathbf{x}}_{0}), \quad i = 1,\dots,n,
$$
produces an independent structure–sequence pair. A scalar reward can then be aggregated as either the mean or maximum across samples,
$$
r^{\text{mean}}({\mathbf{x}}_{0}) = \frac{1}{n}\sum_{i=1}^{n} r(\tilde{\mathbf{x}}^{(i)}_{0}, \mathbf{s}^{(i)}_t), \qquad
r^{\text{max}}({\mathbf{x}}_{0}) = \max_i r(\tilde{\mathbf{x}}^{(i)}_{0}, \mathbf{s}^{(i)}_t).
$$
Because $r$ operates on coherent sequence–structure pairs, any scalar property derived from either the structure or sequence can be incorporated as a reward.

For binder design, the interface quality is quantified by the computed binding free energy $\Delta G$ between the designed and target chains, with
$$
r_{\text{bind}} = -\Delta G.
$$
Sequence-level objectives can also be introduced. To control the electrostatic character, a charge reward penalizes deviation from a target net charge $Q^\ast$ at pH 7,
$$
r_{\text{charge}} = -|Q - Q^\ast|.
$$
Secondary structure composition is constrained through a combined geometry- and sequence-based reward. Structural content is estimated using DSSP on the refined backbone and residue-based propensities from the sequence, combined with fixed weights (0.8 for DSSP and 0.2 for sequence) to yield the helix, $\beta$-sheet, and loop fractions $(\alpha, \beta, \ell)$. The reward penalizes deviation from user-specified targets $(\alpha^\ast, \beta^\ast, \ell^\ast)$,
$$
r_{\text{SS}} = w_{\alpha}(1 - |\alpha - \alpha^\ast|) +
w_{\beta}(1 - |\beta - \beta^\ast|) +
w_{\ell}(1 - |\ell - \ell^\ast|),
$$
where $w_{\alpha}$, $w_{\beta}$, and $w_{\ell}$ control the relative contribution of each term. When steering toward a specific secondary structure, that term is weighted fourfold relative to the others.

\subsection{Steering configuration evaluation}

To characterize the behavior of FK steering during binder design, we performed a systematic configuration sweep targeting streptolysin O, with hotspot residues 110, 115, and 117. Each condition was repeated three times.

To examine the effect of the steering potential, three independent runs were carried out using 50 particles and a binder length ($L_{\text{binder}} = 24$). To investigate the influence of the number of particles ($n_{\text{particles}}$), guidance temperature ($\tau$), resampling interval ($\Delta t$), and guidance onset ($t_{\text{start}}$), we performed three repeats of guided diffusion per parameter, keeping all others fixed at the default values presented in \textbf{Table 2}.

\begin{table}[H]
\centering
\caption{\textbf{Default parameters during the parameter sweep.}}
\begin{tabular}{ll}
\hline
Parameter & Value \\
\hline
$L_{\text{binder}}$ & 15 \\
$\tau$ & 10 \\
$n_{\text{particles}}$ & 20 \\
$t_{\text{start}}$ & 50 \\
$\Delta t$ & 2 \\
$G_t$ & Immediate \\
\hline
\end{tabular}
\end{table}

In each case, only the parameter under evaluation was varied.

\subsection{Assessing the correlation between steerable rewards and designability}

To evaluate the relationship between steerable rewards and designability, we generated 90 designs without resampling for each of the 5 benchmark targets: 5O45, 3DI3, 4ZXB, 1WWW and 5VLI, and re-predicted their complex conformation using Boltz-2. The hotspots defined during designs were: A56, A115, A123 for 5O45; B58, B80, B139 for 3DI3; E64, E88, E96 for 4ZXB; X294, X296, X333 for 1WWW; and B521, B545, B552 for 5VLI.

We then defined designability as the proportion of designs that achieved an RMSD $< 5$ Å to the intended structure and a favorable binding energy upon re-prediction. The interface free energy reward was computed for each design during generation. We then designed binders with FK steering using the interface free energy reward, re-predicted their complex conformations, and compared their designability to unguided binders. During design, we set $n_{\text{particles}}=20$, $\tau=10$, $t_{\text{start}}=20$, $\Delta t=2$, and used the \textit{immediate} potential. We generated 5 sequences per design and used the maximum reward across sequences for steering.

\subsection{Binder design against bacterial virulence factors}

We applied steering to design peptide binders against clinically relevant virulence factors from \textit{Streptococcus pyogenes} and \textit{Streptococcus pneumoniae}. The targets: EndoS (PDB: 8A49), IdeS (PDB: 8A47), pneumolysin (PLY, PDB: 5CR6), C5a peptidase (PDB: 3EIF), and streptolysin O (SLO, PDB: 4HSC), are factors for which neutralizing antibodies have been described and/or have known catalytic sites \cite{ides_endos,c5apeptidase}. These guided the selection of receptor hotspot residues defining the design interfaces.

For SLO, the targeted epitope comprised residues A110, A115, and A117. EndoS was targeted at two antibody-binding regions on chain C: interface 1 spanning residues 765-916 with hotspots C798, C835, and C909, and interface 2 spanning residues 295-421 with hotspots C314, C315, and C316. IdeS was targeted at three distinct regions of chain C: an antibody-binding interface near residues C185-C187, a second interface near C255, C258, and C322, and the catalytic triad formed by C93, C264, and C286. PLY was targeted at an in-house defined epitope (unpublished). C5a peptidase was targeted at the catalytic region, with hotspots A423, A360, A356 and A426.

All campaigns used 50 diffusion particles, resampling every two steps, guidance from timestep 20, and the \textit{immediate} FK potential acting on interface $\Delta G$ reward. Guided and unguided runs were performed for each target. The beta sheet checkpoint was used for RFDiffusion weights.

\section{Code and data availability}
FK-steered RFdiffusion is available at \href{https://github.com/ErikHartman/FK-RFdiffusion}{GitHub}.

\section{Acknowledgements}
We thank Di Tang and Alejandro Gomez Toledo for their insights on target selection.

\printbibliography

\end{document}